\documentclass{article}

\usepackage{graphicx}
\usepackage[main, final]{neurips_2026}
\usepackage[utf8]{inputenc} 
\usepackage[T1]{fontenc}    
\usepackage{hyperref}       
\usepackage{url}            
\usepackage{booktabs}       
\usepackage{amsfonts}       
\usepackage{nicefrac}       
\usepackage{microtype}      
\usepackage{xcolor}         
\usepackage{multirow}
\usepackage{amsmath} 
\usepackage[table]{xcolor}
\usepackage{arydshln}
\usepackage{natbib}
\usepackage{enumitem}
\title{ReCon: Toward Balanced Learning under Inter-Context and Context-Memory Conflicts}

\author{
\makebox[\textwidth][c]{
\begin{tabular}{c}
Huipeng Ma$^{1,2}$ \ 
Dandan Song$^{1,*}$ \ 
Shaonan Ma$^{2,*}$ \ 
Changzhi Zhou$^{1}$ \ 
Jun Yang$^{1}$ \\[1pt]
Yuhang Tian$^{1}$ \ 
Luan Zhang$^{1}$ \ 
Chenhao Li$^{1}$ \ 
Xudong Li$^{1}$ \ 
Fang Xi$^{2}$ \ 
Guangyuan Feng$^{1}$ \\[6pt]
$^{1}$Beijing Institute of Technology \qquad
$^{2}$Qiyuan Lab \\[4pt]
\texttt{\{mahuipeng,sdd\}@bit.edu.cn \quad mashaonan@qiyuanlab.com} 
\end{tabular}
}
}

\begin{document}

\maketitle
\begingroup
\renewcommand{\thefootnote}{*}
\footnotetext{Corresponding author.}
\endgroup

\begin{abstract}
In retrieval-augmented generation (RAG), knowledge conflicts arise when retrieved contexts disagree with each other or with the model's parametric memory. Reinforcement Learning with Verifiable Rewards (RLVR) has emerged as a standard approach to address such conflicts. 
However, existing RLVR-based methods are typically formulated under a single-conflict assumption.
In realistic mixed-conflict settings where Inter-Context and Context-Memory co-exist, unified RLVR optimization does not distinguish conflict types. 
As a result, the type with stronger advantage signals dominates policy updates, leaving the other under-optimized and thus yielding imbalanced performance across conflict types.
To address this challenge, we propose \textbf{ReCon}, a type-aware RLVR framework for balanced learning across Inter-Context and Context-Memory conflicts.
Specifically, ReCon computes a step-level imbalance signal from type-wise mean absolute advantage and triggers resampling on the weaker conflict type. 
Instead of resampling from scratch, ReCon branches from informative failures.
It selects an informative failed trajectory using uncertainty and closeness signals, then branches at an entropy-jump point, concentrating additional rollouts on critical reasoning decisions.
Experiments show that ReCon yields stronger and more balanced performance across knowledge-conflict and multi-hop QA benchmarks, and also generalizing well under mixed-conflict settings.

\end{abstract}

\section{Introduction}
Retrieval-augmented generation (RAG) introduces external contexts to large language models during reasoning, enabling them to exploit retrieved evidence in addition to their parametric knowledge~\cite{borgeaud2022improving, REALM, FiD, lewis2020retrieval,  Ma_Zhang_Song_Hu_Tian_Yang_Zhou_Li_Jin_Li_Lin_Zhang_Zhang_2026,  zhang2023iag, zhu2026logictree}. Despite this, introducing external evidence forces the model to confront multiple sources of information that may conflict during reasoning. Making reliable decisions when confronted with such conflicting evidence has become one of the central challenges in understanding and improving the behavior of RAG models~\cite{chen2022rich, xu2024knowledge}.  

Prior work has proposed various methods to handle knowledge conflicts. Several methods focus on Context-Memory conflicts, aiming to reconcile discrepancies between retrieved evidence and the model’s internal knowledge\cite{bi2025parameters, gao2023rarr, wang2025adacad}. Others investigate Inter-Context conflicts, studying how to mitigate disagreements across multiple retrieved contexts~\cite{chen2022rich, su2024conflictbank, wang2025retrieval}. 

With the success of DeepSeek-R1~\cite{guo2025deepseek, lambert2024tulu}, Reinforcement Learning with Verifiable Rewards (RLVR)~\cite{lambert2024tulu, drgrpo, dapo, gspo} has emerged as an effective post-training paradigm for improving the reasoning abilities of LLMs. Recent studies have also applied RLVR to knowledge-conflict settings and use verifiable feedback to model different conflict types in a targeted manner. For Inter-Context conflicts, KCR~\cite{zheng2025kcr} pairs contradictory contexts and optimizes verifiable rewards to encourage coherent selection. For Context-Memory conflicts, Trustworthy-Alignment~\cite{zhang2024trustworthy} uses counterfactual contexts to align the model toward context-grounded answers, while Knowledgeable-R1~\cite{lin2025knowledgeable} leverages multiple specialized policies to decide when to follow context or rely on parametric knowledge.

However, these RLVR-based approaches are typically formulated under a single-conflict assumption. In practice, models often need to handle mixed-conflict scenarios where both conflict types co-exist. We therefore consider a more realistic mixed-conflict setting, where Inter-Context and Context-Memory conflicts co-exist and require joint optimization. 
Nevertheless, under RLVR optimization, naively mixing both conflict types in the training data is not sufficient.
These RLVR pipelines typically apply the same advantage estimator to all conflict examples and aggregating them into a single update step.
As a result, the training process does not explicitly account for differences in characteristics or difficulty between different types of conflicts. In this uniform treatment, conflict types that contribute more substantial advantage signals progressively dominate policy updates, whereas those with persistently weaker advantages are diluted during gradient aggregation and remain under-optimized, leading to a systematic performance imbalance across conflict types.

Motivated by this, we propose \textbf{ReCon}, which rebalances type-wise mean absolute advantages for balanced learning under mixed knowledge conflicts.
Specifically, it first partitions training examples into Inter-Context and Context-Memory groups, and tracks the mean absolute advantage of each group during training. When a clear imbalance in type-wise mean absolute advantages is detected, ReCon treats the conflict type with the smaller mean absolute advantage as the weaker type. It then preferentially resamples examples from this type to amplify its effective learning signal and promote more balanced policy optimization across conflict types. Rather than resampling from scratch, this resampling is implemented by selecting an informative failed trajectory using uncertainty and closeness signals, with subsequent branching at an entropy-jump point along it, thereby allocating additional rollouts to critical reasoning decisions.


Experiments on four conflict and multi-hop QA benchmarks (ConflictBANK, RAMDocs, HotpotQA, 2WikiMultiHopQA), across three backbones spanning different scales and families (Qwen2.5-3B/7B-Instruct and Llama-3.1-8B-Instruct), show that ReCon yields stronger and more balanced performance over strong baselines, while generalizing well under mixed-conflict settings. Our contributions can be summarized as follows:
\begin{itemize}[itemsep=2pt,topsep=0pt,parsep=0pt,leftmargin=*]
    \item We introduce ReCon for balanced learning under mixed knowledge conflicts, explicitly addressing the optimization imbalance between Inter-Context and Context-Memory conflicts.
    \item We design a sampling strategy within ReCon that resamples informative examples from the weaker conflict type to rebalance policy improvement.
    \item We conduct experiments on knowledge-conflict and multi-hop QA benchmarks, showing that ReCon yields stronger and more balanced performance across mixed-conflict settings.
\end{itemize}

\section{Preliminary}
\label{gen_inst}
\subsection{Knowledge Conflict.}
We consider context-conditioned question answering. Each sample is represented as
$(q,\mathcal{C},a^{\ast})$, where $q$ is the input question,
$\mathcal{C}=\{c_1,\ldots,c_m\}$ is the associated context set, and $a^{\ast}$ is the target answer.
Different $c_i$ may correspond to different candidate answers, thereby allowing $\mathcal{C}$ to contain
multiple mutually inconsistent candidate answers (Inter-Context conflicts).
Moreover, the conclusion implied by $\mathcal{C}$ may contradict the model's parametric memory
(Context-Memory conflicts).
Given $(q,\mathcal{C})$ and a prompt template $p$,
the model $\pi_{\theta}$ generates an answer
$y \sim \pi_{\theta}(\cdot \mid q,\mathcal{C},p)$, and correctness is measured by the agreement between
$y$ and $a^{\ast}$.

\subsection{RLVR and GRPO.}

We follow the RLVR paradigm, which trains LLMs with rewards derived from deterministic verifiers rather than learned reward models. Given a generated answer $y$ and a ground-truth answer $a^{\ast}$, a verifier $V(\cdot)$ returns a scalar reward $r = V(y, a^{\ast})$ that reflects answer correctness. 

For policy optimization, we employ GRPO, which removes the value model by replacing it with group-wise reward normalization and a KL-regularized clipped policy gradient. For an instance $(q, \mathcal{C}, a^{\ast})$, the policy $\pi_{\theta}$ generates $G$ answers $\{y_i\}_{i=1}^{G}$ with verifier rewards $\{r_i\}_{i=1}^{G}$. Let $\mu_r$ and $\sigma_r$ denote the group mean and standard deviation of $\{r_i\}$. GRPO computes the group-relative advantage
\begin{equation}
\hat{A}_i = (r_i - \mu_r) / (\sigma_r + \varepsilon),
\end{equation}
where $\varepsilon > 0$ prevents division by zero. We define the importance ratio $\rho_i = \pi_{\theta}(y_i \mid q, \mathcal{C}) / \pi_{\text{old}}(y_i \mid q, \mathcal{C})$, and denote $\text{CLIP} = \text{clip}(\rho_i, 1-\epsilon, 1+\epsilon)$. The GRPO objective is
\begin{equation}
\begin{aligned}
\mathcal{L}_{\text{GRPO}}(\theta)
=&-\mathbb{E}_{i}\Big[\min\big(\rho_i\hat{A}_i,\ \text{CLIP}\cdot \hat{A}_i\big)\Big]
+\beta\,\mathbb{E}_{i}\Big[\mathrm{KL}\big(\pi_{\theta}\ \|\ \pi_{\text{ref}}\big)\Big],
\end{aligned}
\end{equation}
where $\pi_{\text{ref}}$ is a fixed reference policy and $\beta$ controls the KL regularization strength. Throughout the paper, we instantiate the verifier $V$ as an exact-match check between $y$ and $a^{\ast}$, so $r_i \in \{0, 1\}$.


\begin{figure*}[t]
\centering
\includegraphics[width=0.85\textwidth]{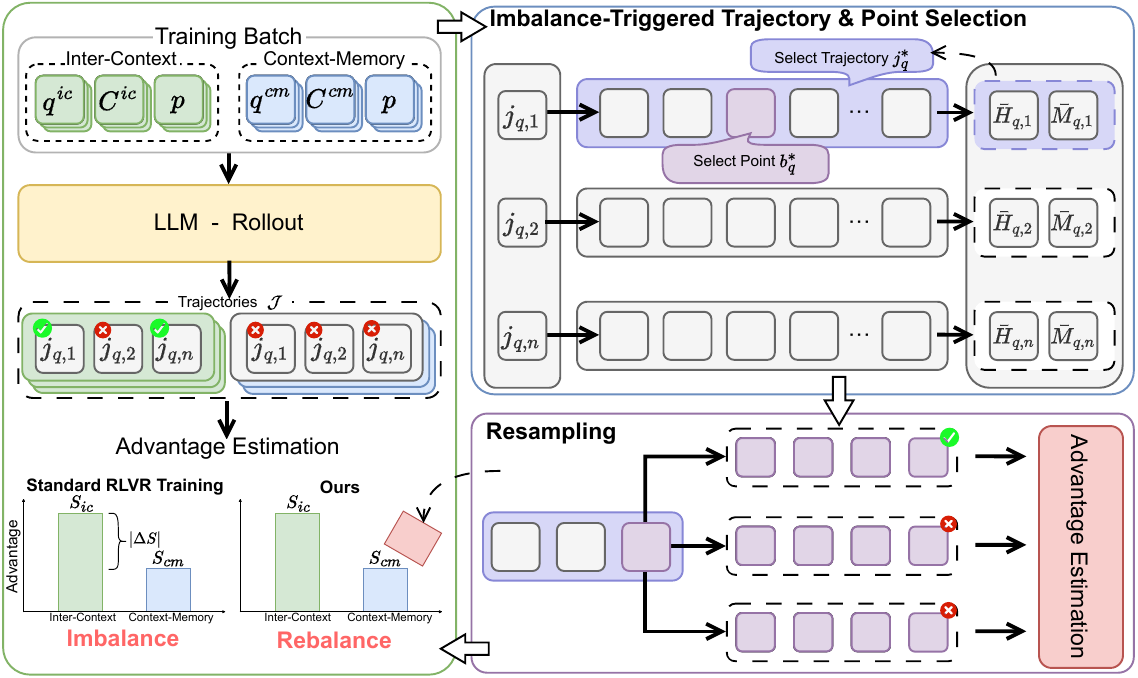} 
\caption{Overview of ReCon. Upon detecting type imbalance, it selects a branching trajectory and point, then resamples from that point to generate additional rollouts and update the policy.
} 
\label{fig:framework}
\end{figure*}

\section{Method}
\label{headings}
We propose \textbf{ReCon}, which incorporates Inter-Context and Context-Memory knowledge conflicts in a unified modeling and optimization process.
As shown in Figure~\ref{fig:framework}, ReCon continuously monitors type-wise mean absolute advantages during training to detect imbalance across conflict types. Once imbalance is observed, it selectively resamples more informative instances from the weaker type to rebalance policy improvement.

\subsection{Step-level Imbalance Signal}
\label{sec:method-tca}

We introduce a step-level statistic that quantifies the update 
strength contributed by each conflict type. Since GRPO scales 
policy updates by advantages, the mean absolute advantage over 
a subset of samples directly reflects the effective learning 
signal from that subset.

\paragraph{Type-wise mean absolute advantage.}
Let $\mathcal{T}=\{\mathrm{ic},\mathrm{cm}\}$ denote the two 
conflict types. At each training step, we sample $G$ rollouts 
per question in the current mini-batch, and let $A_{q,k}$ 
denote the GRPO advantage of the $k$-th rollout for question $q$ 
with conflict-type label $t_q$. The type-wise cumulative 
advantage is
\begin{equation}
S_t \;=\; \frac{1}{G\,|\mathcal{Q}_t|}
\sum_{q\in\mathcal{Q}_t}\sum_{k=1}^{G}\big|A_{q,k}\big|,
\qquad t\in\mathcal{T}.
\end{equation}
where $\mathcal{Q}_t=\{q:t_q=t\}$ is the set of questions of type $t$ in 
the current mini-batch. 

\paragraph{Imbalance-triggered weaker type.}
We define the step-level imbalance as 
$\Delta S = S_{\mathrm{ic}} - S_{\mathrm{cm}}$, 
and identify the weaker conflict type as the one with the 
smaller mean absolute advantage:
\begin{equation}
t^{\ast} \;=\; \arg\min_{t\in\mathcal{T}} S_t.
\end{equation}
Resampling is triggered only when the imbalance is sufficiently 
pronounced, i.e., $|\Delta S| > \epsilon_{\text{stop}}$, where 
$\epsilon_{\text{stop}} > 0$ is a predefined threshold.





\subsection{Imbalance-Triggered Resampling}
\label{sec:method-resample}

Once imbalance is detected and the weaker type $t^{\ast}$ is 
identified (Section~\ref{sec:method-tca}), ReCon increases 
exploration on $t^{\ast}$ by selectively resampling informative 
questions. Naive resampling from scratch, however, is of limited 
benefit when all rollouts fail. ReCon instead focuses the extra 
compute on the most informative part of the reasoning process, 
by branching from a high-uncertainty failed trajectory at a 
critical decision point that is more likely to alter the final 
outcome.

\subsubsection{Failure buffer maintenance.}

We maintain a failure buffer $\mathcal{F}_{t^{\ast}}$ of recent 
failed questions for the weaker type. A question $q$ is inserted 
into $\mathcal{F}_{t^{\ast}}$ when all $G$ rollouts fail the 
verifier in the current step, and is removed once it yields at 
least one successful rollout. To retain only recent failures, 
$\mathcal{F}_{t^{\ast}}$ is implemented as a sliding window of 
size $W$.

When resampling is triggered, ReCon uniformly samples $R$ 
candidate questions from $\mathcal{F}_{t^{\ast}}$; if the buffer 
is empty, resampling is skipped for this step.

\subsubsection{Trajectory selection.}

For each candidate question $q$, we select a failed trajectory 
$j_q^{\ast}$ from $\mathcal{J}_q=\{j_{q,k}\}_{k=1}^{G}$ based on 
two complementary signals: an \emph{entropy-based uncertainty} 
signal that captures where the policy is unsure, and a 
\emph{preference-based closeness} signal that captures whether 
the policy is close to recovering the ground-truth answer.

\paragraph{Entropy-based uncertainty signal.}
A failed trajectory that exposes \emph{critical} reasoning 
decisions is more likely to yield informative branches when 
resampled. We capture this with the mean token-level entropy 
along the trajectory,
\begin{equation}
\bar{H}_{q,k} 
\;=\; \mathrm{Norm}_q\!\left(
\frac{1}{L}\sum_{u=1}^{L} H_{q,k}(u)
\right),
\end{equation}
where $H_{q,k}(u)$ is the token entropy at position $u$ of 
trajectory $j_{q,k}$, $L = |j_{q,k}|$ is its length in tokens, 
and $\mathrm{Norm}_q(\cdot)$ denotes min-max normalization 
across the $G$ trajectories of question $q$.

\paragraph{Preference-based closeness signal.}
Entropy alone reflects uncertainty, but not whether the 
probability mass is actually moving toward the ground-truth 
answer. We therefore complement entropy with the policy's 
\emph{relative preference} for the ground-truth answer over the 
generated one, evaluated under the same teacher-forcing prefix 
$\mathrm{pref}(q,k)$, i.e., the reasoning segment of trajectory 
$j_{q,k}$ that precedes the final answer. The closeness score is
\begin{equation}
\bar{M}_{q,k} 
\;=\; 1 - \mathrm{Norm}_q\!\left(
\ell_\theta(a_q^{\ast}) - \ell_\theta(\tilde{a}_{q,k})
\right),
\end{equation}
where $\ell_\theta(a) = \tfrac{1}{|a|}\log p_\theta(a \mid q, 
\mathcal{C}, \mathrm{pref}(q,k))$ is the length-normalized 
log-probability of answer $a$, $a_q^{\ast}$ is the ground-truth 
answer, and $\tilde{a}_{q,k}$ is the answer produced by 
trajectory $j_{q,k}$.

\paragraph{Trajectory selection.}
We combine the two signals into a joint score and select the 
trajectory with the highest score,
\begin{equation}
j_q^{\ast} 
\;=\; \arg\max_{k\in\{1,\ldots,G\}}
\left[\,
\alpha_e\,\bar{H}_{q,k} + (1-\alpha_e)\,\bar{M}_{q,k}
\,\right],
\end{equation}
where $\alpha_e \in [0,1]$ balances uncertainty and closeness.

\subsubsection{Branching point selection.}

Given the token-level entropy sequence 
$\{H_{q,\ast}(u)\}_{u=1}^{L_{q,\ast}}$ along the selected 
trajectory $j_q^{\ast}$, we branch at a position where 
uncertainty rises sharply, which often corresponds to a 
critical reasoning decision. We first take the top-$K$ 
highest-entropy positions as candidates, denoted by 
$\mathcal{O}_K$. For each $s\in\mathcal{O}_K$, we compute a 
local entropy-jump score against its preceding context, and 
select the branching point with the largest jump,
\begin{equation}
b_q^{\ast} 
\;=\; \arg\max_{s\in\mathcal{O}_K}\left[
H_{q,\ast}(s) - \frac{1}{\ell_s}\sum_{u=s-\ell_s}^{s-1} H_{q,\ast}(u)
\right],
\end{equation}
where $\ell_s = \min(w, s-1)$ is the length of the look-back 
window and $w$ is a fixed window size.

\subsection{Policy Optimization}
\label{sec:method-pipeline}
At each RLVR step, we sample $G$ rollouts per instance and compute GRPO advantages from verifier rewards. A verifiable correctness reward based on ACC is used, assigning a reward of 1 when the predicted answer matches the gold answer and 0 otherwise. We then aggregate advantages by conflict type to obtain $S_{\mathrm{ic}}, S_{\mathrm{cm}}$
and an imbalance signal $\Delta S$; if $|\Delta S|$ is small we proceed with a standard GRPO update, otherwise we identify the weaker type $t^{\ast}$ and trigger resampling. When resampling is triggered, we draw candidate questions from the failure buffer $\mathcal{F}_{t^{\ast}}$. For each candidate, ReCon branches from an informative failed trajectory selected by entropy and closeness signals, chooses a branching point by an entropy-jump criterion, and samples extra rollouts to augment the current-step data before performing the GRPO update.

\section{Experimental Setup}
\label{es}

\subsection{Datasets}
We evaluate ReCon in two settings. For knowledge-conflict benchmarks, we train on a subset of \textbf{ConflictBANK}~\cite{su2024conflictbank} and evaluate on both \textbf{ConflictBANK} and \textbf{RAMDocs}~\cite{wang2025retrieval}. ConflictBANK provides each question with supporting and conflicting evidence documents, making it suitable for constructing mixed-conflict training data. Based on it, we build two conflict settings, \textit{Inter-Context} and \textit{Context-Memory}, as described in Appendix~\ref{conflictbank}. 
We use RAMDocs as an additional benchmark to assess cross-benchmark generalization under more diverse sources of conflicting evidence. 
For multi-hop QA benchmarks, we further construct counterfactual passages to explicitly introduce knowledge conflicts. We train ReCon on a subset of \textbf{HotpotQA}~\cite{hotpotqa} and evaluate on the validation sets of \textbf{HotpotQA} and \textbf{2WikiMultiHopQA}~\cite{2wikimqa}. Details of the counterfactual construction are provided in Appendix~\ref{hotpotqa}.


\subsection{Implementation Details}
All experiments are implemented in the \textsc{VERL}~\cite{verl} framework with GRPO~\cite{guo2025deepseek}. We use Qwen2.5-3B-Instruct, Qwen2.5-7B-Instruct~\cite{team2024qwen2}, and Llama-3.1-8B-Instruct~\cite{grattafiori2024llama} as the backbones and run training on 8$\times$ NVIDIA A800 GPUs. We set the batch size to 256, max prompt to 2048, response lengths to 512, and optimize with learning rate $1\times10^{-6}$. For rollouts, we sample 5 responses with temperature 0.7, top-$k$ 10, and top-$p$ 0.95. We report micro-F1 score as the evaluation metric, which measures the overlap between the generated answer and the ground truth. Prompt templates are provided in Appendix~\ref{prompttmp}. 

\subsection{Baselines}
We compare ReCon with several representative baselines. \textbf{CK-PLUG}~\cite{bi2025parameters} is a decoding-time method that detects knowledge conflicts via confidence gain and adjusts token probabilities to control parametric versus contextual reliance. \textbf{GRPO}~\cite{guo2025deepseek} is a verifier-based RL baseline with group-wise reward normalization. \textbf{Trustworthy-Alignment}~\cite{zhang2024trustworthy} and \textbf{Knowledgeable-R1}~\cite{lin2025knowledgeable} are two representative methods for Context-Memory conflicts, respectively encouraging context-grounded answers and stronger reliance on parametric knowledge under misleading evidence. \textbf{KCR}~\cite{zheng2025kcr} is an Inter-Context conflict baseline that organizes conflicting contexts into reasoning paths and optimizes logic-based rewards. We further include \textbf{SeqGRPO}, a sequential training variant of GRPO inspired by curriculum-style optimization~\cite{cs3, cs1, cs2}, which trains on one conflict type first and then the other. We evaluate both orders, IC$\rightarrow$CM and CM$\rightarrow$IC, to test whether mixed conflict types can be handled effectively through sequential optimization.


\begin{table*}[t]
\centering
\scriptsize
\caption{Main results (F1) on knowledge-conflict benchmarks and multi-hop QA benchmarks. For ConflictBANK and RAMDocs, IC and CM denote Inter-Context and Context-Memory conflicts, respectively, and ALL denotes the aggregate performance across the two conflict types. Best results within each model block are in bold.}
\begin{tabular}{lcccccccccc}
\toprule
\multirow{2}{*}{\textbf{Method}}
& \multicolumn{3}{c}{\textbf{ConflictBANK}}
& & \multicolumn{3}{c}{\textbf{RAMDocs}} 
& & \multicolumn{2}{c}{\textbf{Multi-hop QA}} \\
\cmidrule{2-4} \cmidrule{6-8} \cmidrule{10-11}
& \textbf{IC} & \textbf{CM} & \textbf{ALL}
& & \textbf{IC} & \textbf{CM} & \textbf{ALL} & & \textbf{HotpotQA} & \textbf{2WikiMQA} \\
\midrule
\textbf{Qwen2.5-3B-Instruct} \\
CK-PLUG & 51.20 & 31.48 & 40.29 & & 42.11 & 39.25 & 40.74 & & 13.43 & 14.98 \\
GRPO & 68.70 & 54.62 & 64.97 & & 50.86 & 36.00 & 44.92 & & 20.44 & 23.82 \\
Trustworthy-Alignment & 63.79 & 50.03 & 55.24 & & 41.53 & 30.79 & 35.11 & & 22.75 & 26.21 \\
Knowledgeable-R1 & 63.24 & 57.11 & 60.17 & & 46.83 & 39.28 & 42.69 & & 26.96 & 29.63 \\
KCR & 73.07 & 53.21 & 66.17 & & 52.67 & 32.78 & 43.11 & & 29.97 & 32.44 \\
SeqGRPO (IC$\rightarrow$CM) & 61.90 & 56.00 & 58.14 & & 43.54 & 35.33 & 38.86 & & 20.16 & 25.64 \\
SeqGRPO (CM$\rightarrow$IC) & 69.29 & 51.39 & 61.46 & & 50.85 & 34.73 & 43.01 & & 23.77 & 24.67 \\
\rowcolor{blue!10} ReCon (Ours) & \textbf{75.11} & \textbf{63.45} & \textbf{71.96} & & \textbf{54.72} & \textbf{44.93} & \textbf{51.78} & & \textbf{31.67} & \textbf{34.21} \\
\hline
\textbf{Qwen2.5-7B-Instruct} \\
CK-PLUG & 63.23 & 47.70 & 56.11 & & 42.33 & 39.05 & 40.54 & &  18.74 & 19.76 \\
GRPO & 78.43 & 74.15 & 76.59 & & 57.19 & 49.21 & 53.27 & & 26.85 & 29.63 \\
Trustworthy-Alignment & 77.84 & 73.27 & 75.98 & & 51.00 & 44.10 & 47.64 & & 29.65 & 32.22 \\
Knowledgeable-R1 & 74.83 & 78.24 & 76.35 & & 56.60 & 52.27 & 54.26 & & 31.54 & 31.43 \\
KCR & \textbf{82.20} & 74.60 & 79.33 & & \textbf{59.35} & 46.11 & 52.31 & & 35.17 & 34.96 \\
SeqGRPO (IC$\rightarrow$CM) & 70.76 & 77.33 & 73.15 & & 49.52 & 50.17 & 49.92 & & 25.11 & 30.73 \\
SeqGRPO (CM$\rightarrow$IC) & 79.74 & 69.24 & 75.09 & & 58.21 & 47.65 & 52.66 & & 28.77 & 31.13 \\
\rowcolor{blue!10} ReCon (Ours) & 82.03 & \textbf{82.97} & \textbf{82.75} & & 59.21 & \textbf{57.97} & \textbf{58.74} & & \textbf{37.67} & \textbf{40.18} \\
\hline
\textbf{Llama-3.1-8B-Instruct} \\
CK-PLUG                                  & 59.71 & 42.26 & 51.35 & & 44.98 & 41.17 & 42.96 & & 18.57 & 20.62 \\
GRPO                                     & 74.36 & 70.45 & 72.59 & & 53.79 & 48.06 & 51.73 & & 25.15 & 27.87 \\
Trustworthy-Alignment                    & 76.93 & 69.41 & 73.51 & & 53.01 & 43.65 & 48.82 & & 32.65 & 35.13 \\
Knowledgeable-R1                         & 67.71 & \textbf{77.92} & 72.84 & & 50.47 & 53.14 & 52.59 & & 31.11 & 32.75 \\
KCR                                      & 78.04 & 69.71 & 74.29 & & 57.27 & 44.81 & 50.04 & & 33.67 & 35.59 \\
SeqGRPO (IC$\rightarrow$CM)              & 69.95 & 74.13 & 72.19 & & 50.01 & 48.56 & 49.86 & & 24.97 & 28.78 \\
SeqGRPO (CM$\rightarrow$IC)              & 74.18 & 67.55 & 70.83 & & 52.00 & 46.58 & 49.03 & & 26.76 & 30.83 \\
\rowcolor{blue!10} ReCon (Ours)         & \textbf{79.69} & 76.97 & \textbf{77.03} & & \textbf{58.57} & \textbf{54.14} & \textbf{56.66} & & \textbf{35.26} & \textbf{38.35} \\
\bottomrule
\end{tabular}
\label{tab:main_results}
\end{table*}

\section{Experimental Results}
\label{exper}
\subsection{Main Results}
The experimental results are presented in Table~\ref{tab:main_results}. We can obtain the following observations:
\paragraph{Competitive performance under mixed conflicts.}
Table~\ref{tab:main_results} reports F1 results on ConflictBANK, RAMDocs, and two multi-hop QA benchmarks across Qwen2.5-3B-Instruct, Qwen2.5-7B-Instruct, and Llama-3.1-8B-Instruct. On ConflictBANK, ReCon consistently achieves the best overall performance across all three model scales. More importantly, the gains are consistent at the type level: ReCon improves both Inter-Context (IC) and Context-Memory (CM) scores rather than overfitting to a single type. In particular, ReCon remains competitive with the strongest type-specialized baseline on each conflict type (KCR on IC, Knowledgeable-R1 on CM), while consistently achieving the best overall score (ALL) across all three backbones.

\paragraph{Generalization across backbones and benchmarks.}
Beyond in-domain performance on ConflictBANK, ReCon also generalizes well across backbones and benchmarks. 
Across three backbones spanning different scales and model families (Qwen2.5-3B-Instruct, Qwen2.5-7B-Instruct, and Llama-3.1-8B-Instruct), ReCon consistently achieves the best overall performance, indicating that its type-aware rebalancing does not rely on a specific model size or pretraining distribution. On RAMDocs, ReCon (trained on ConflictBANK) maintains the best overall F1 with notable gains on CM, demonstrating cross-benchmark robustness under more diverse sources of conflicting evidence. A similar trend holds on the multi-hop QA benchmarks, where ReCon consistently outperforms strong baselines on both in-domain HotpotQA and the out-of-distribution 2WikiMQA. 

\paragraph{Balanced improvements across conflict types.}
Table~\ref{tab:main_results} reveals a clear imbalance pattern among existing type-specialized methods. 
KCR, which targets IC conflicts, excels on IC but degrades noticeably on CM; in contrast, Trustworthy-Alignment and Knowledgeable-R1, both designed for CM conflicts, exhibit the opposite trend. This indicates that type-specialized optimization only amplifies the model's response to one conflict signal, while the learning signal of the other conflict type is systematically suppressed, leading to a structural type bias. In contrast, ReCon does not optimize for a single conflict type; instead, it continuously monitors the type-wise mean absolute advantages of both conflicts during training and triggers resampling on the weaker type once imbalance is detected, thereby allocating additional rollout budget where it is most needed. This mechanism-level rebalancing preserves the learning signals of both IC and CM conflicts, yielding stable improvements on both types simultaneously rather than trading one off against the other.

\paragraph{Sequential training causes type forgetting.}
A natural question is whether IC and CM conflicts can be addressed via curriculum-style sequential training, where the model is first trained on one conflict type until convergence and then on the other. The two variants of SeqGRPO (IC$\rightarrow$CM and CM$\rightarrow$IC) directly test this hypothesis. As shown in Table~\ref{tab:main_results}, regardless of the training order, the conflict type trained in the first stage consistently suffers a clear performance drop in the second stage, even falling below the single-stage GRPO baseline. This indicates that IC and CM are not related by difficulty level but exhibit a degree of objective interference, analogous to catastrophic forgetting in multi-task learning. This finding directly motivates the central design of ReCon: the two conflict types must be jointly rebalanced within a single optimization process rather than handled sequentially. 
ReCon substantially outperforms both SeqGRPO variants, showing that type-aware online rebalancing handles mixed-conflict scenarios more effectively than any static training schedule.
\begin{table}[t]
\centering
\scriptsize
\caption{Ablation results on ConflictBANK, RAMDocs, HotpotQA, and 2WikiMQA using Qwen2.5-7B-Instruct, measured by F1. The upper section shows the effect of resampling, the middle section shows trajectory selection ablations, and the lower section shows branching point selection ablations.}
\begin{tabular}{lcccccccccc}
\toprule
\multirow{2}{*}{\textbf{Method}}
& \multicolumn{3}{c}{\textbf{ConflictBANK}}
& & \multicolumn{3}{c}{\textbf{RAMDocs}} 
& & \multicolumn{2}{c}{\textbf{Multi-hop QA}} \\
\cmidrule{2-4} \cmidrule{6-8} \cmidrule{10-11}
& \textbf{IC} & \textbf{CM} & \textbf{ALL}
& & \textbf{IC} & \textbf{CM} & \textbf{ALL} & & \textbf{HotpotQA} & \textbf{2WikiMQA} \\
\midrule
GRPO & 78.43 & 74.15 & 76.59 & & 57.19 & 49.21 & 53.27 & & 26.85 & 29.63 \\
\rowcolor{blue!10} ReCon (Ours) & \textbf{82.03} & \textbf{82.97} & \textbf{82.75} & & \textbf{59.21} & \textbf{57.97} & \textbf{58.74} & & \textbf{37.67} & \textbf{40.18} \\
\hdashline
\ \ -\textit{w} scratch & 79.16 & 76.33 & 77.52 & & 57.74 & 51.38 & 54.56 & & 27.02 & 29.76 \\
\ \ -\textit{w} type-agnostic & \textbf{82.32} & 74.21 & 77.83 & & -- & -- & -- & & -- & -- \\
\hdashline
\ \ -\textit{w/o} uncertainty & 81.94 & 79.37 & 80.92 & & 58.24 & 55.78 & 57.03 & & 35.45 & 38.31 \\
\ \ -\textit{w/o} closeness & 79.94 & 77.36 & 78.00 & & 57.86 & 52.00 & 55.21 & & 29.17 & 32.74 \\
\hdashline
\ \ -\textit{w} random & 79.98 & 78.36 & 78.73 & & 58.00 & 52.79 & 55.34 & & 30.06 & 33.91 \\
\ \ -\textit{w} top-entropy & 80.11 & 79.37 & 79.68 & & 57.08 & 55.16 & 56.25 & & 33.83 & 35.88 \\
\bottomrule
\end{tabular}
\label{tab:ablation}
\end{table}

\subsection{Analysis of Resampling}
We analyze the effect of resampling by comparing ReCon with an alternative from-scratch resampling approach.
After identifying the weaker conflict type, this approach increases the rollout compute by drawing additional full rollouts from scratch under the same input, without any trajectory- or point-level guidance.
As shown in the upper part of Table~\ref{tab:ablation}, from-scratch resampling yields limited improvements on the weaker type.
A key reason is that, in the all-failure regime, simply drawing more full rollouts from the initial prompt has a low probability of reaching the correct answer.
By contrast, ReCon branches from informative failures and concentrates additional rollouts around critical decisions, which more directly increases the chance of correcting the reasoning and produces stronger learning signals for the weaker type.

We also evaluate a type-agnostic resampler. It uses the same resampling procedure as ReCon but draws failures from a unified IC--CM buffer without identifying the weaker type. It therefore resamples at every step, with the per-step budget adjusted so that the total number of additional rollouts matches ReCon. On ConflictBANK, this variant is comparable to ReCon on IC but falls far behind on CM, widening the IC--CM gap from 0.94 to 8.11 points. A likely reason is that the model is already stronger on IC, so IC failures are easier to repair, and type-agnostic branching keeps reinforcing IC while leaving CM under-optimized. This shows that the gains of ReCon do not come from additional branching alone; deciding which conflict type to explore is also essential.

\begin{figure}[t]
\centering
\includegraphics[width=0.95\columnwidth]{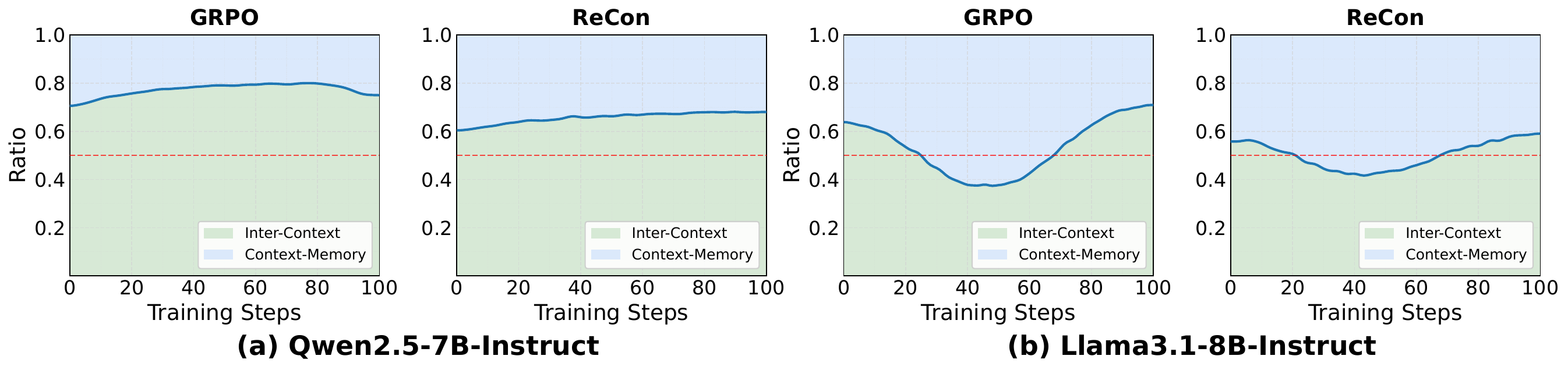}
\caption{Type-wise cumulative-advantage ratio over training steps on ConflictBANK. The red dashed line ($0.5$) marks perfect balance between the two conflict types.}

\label{gvk}
\end{figure}

We further examine how optimization effort is distributed across conflict types during training. At each step, we measure the IC ratio $S_{\mathrm{ic}}/(S_{\mathrm{ic}}+S_{\mathrm{cm}})$ and visualize it over training for both GRPO and ReCon on Qwen2.5-7B-Instruct and Llama-3.1-8B-Instruct.
As shown in Figure~\ref{gvk}, GRPO exhibits clear type-wise imbalance throughout training, but the manifestation differs across backbones: on Qwen2.5-7B-Instruct the ratio remains stably above $0.5$ and gradually drifts further from balance, while on Llama-3.1-8B-Instruct it varies substantially, dropping below $0.5$ before rebounding. In contrast, ReCon maintains a markedly more balanced and stable ratio across both backbones, staying closer to the $0.5$ reference with much less variation. 
These results support our motivation that, without type-aware control, mixed-conflict optimization is systematically biased toward whichever conflict type produces stronger advantage signals, manifesting as either persistent dominance or unstable oscillation depending on the backbone. ReCon's imbalance-triggered resampling effectively mitigates both failure modes in a sustained manner.

\subsection{Analysis of Trajectory Selection}
\begin{figure}[t]
\centering
\includegraphics[width=0.8\linewidth]{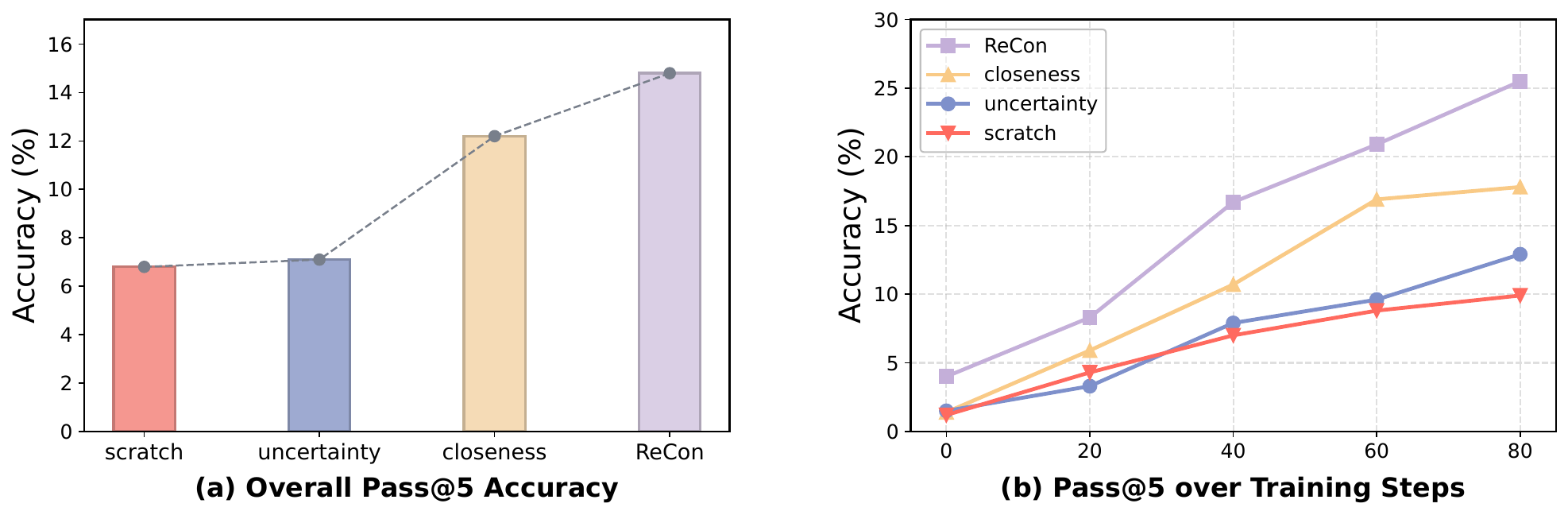}
\caption{Pass@5 accuracy on ConflictBANK with Qwen2.5-7B-Instruct for scratch, uncertainty-only, closeness-only, and ReCon. (a) Overall Pass@5 accuracy. (b) Pass@5 over training steps.}
\label{bar}
\end{figure}

We conduct an ablation study to analyze trajectory selection in ReCon. 
We examine how an entropy-based uncertainty signal interacts with a preference-based closeness signal.
As shown in the middle part of Table~\ref{tab:ablation}, ReCon achieves the strongest and most consistent improvements when both signals are enabled, suggesting that the two signals work well together. The ablations also show that the two signals do not contribute equally. Using the preference-based closeness signal alone still retains non-trivial gains. This suggests that selecting trajectories closer to the desired answer behavior is already a strong criterion for effective correction.
In contrast, using the entropy-based uncertainty signal alone yields smaller gains than using the preference-based closeness signal alone, suggesting that uncertainty by itself is a weaker indicator of which trajectories are correctable under mixed conflicts. These results suggest a clear synergy between the two signals. Closeness provides the primary guidance on which trajectory to fix, while entropy acts as an auxiliary cue that refines where to intervene when combined with closeness. Together, they enable more efficient and sustained rebalancing.
To corroborate the above ablation findings from a sampling-centric perspective, we report Pass@K accuracy on ConflictBANK with Qwen2.5-7B-Instruct. As shown in Figure~\ref{bar}(a), ReCon achieves the highest overall Pass@5 among all ablated settings, showing that its trajectory selection strategy translates into more successful sampling outcomes. Figure~\ref{bar}(b) further tracks Pass@5 as training progresses. While all methods improve over time, ReCon rises more rapidly and maintains a widening margin in later steps. Notably, closeness-only grows faster than uncertainty-only, while scratch remains consistently lowest, mirroring the ranking observed in the ablation table. Together, these curves suggest that ReCon's advantage is not merely an end-point effect; rather, it consistently improves the probability of sampling a correct trajectory during training, indicating a more effective use of the rollout budget.



\subsection{Analysis of Branching Point Selection}
We then investigate the impact of different branching point selection strategies in ReCon. After selecting an informative failed trajectory, ReCon branches at an entropy-jump point. We compare this strategy with two alternatives: \textit{random}, which branches from a randomly chosen point on the selected trajectory, and \textit{top-entropy}, which branches from the point with the highest token entropy. As shown in the bottom part of Table~\ref{tab:ablation}, both alternatives perform worse than the full method. Random branching often selects positions unrelated to the key reasoning error, leading to limited corrective benefit. Branching at the highest-entropy point performs better, indicating that uncertainty is a useful signal for intervention. However, it still underperforms ReCon, suggesting that high entropy alone is insufficient for identifying the most effective branching location. We attribute this to the fact that the highest-entropy point may reflect local ambiguity rather than a true reasoning transition. In contrast, the entropy-jump criterion used in ReCon favors positions where uncertainty rises sharply relative to the preceding context, making them more likely to correspond to critical reasoning decisions. This allows ReCon to allocate additional rollouts more effectively and achieve stronger gains.

\section{Related Work}
\subsection{Knowledge Conflicts}
\paragraph{Knowledge Conflict in LLMs.} Recent work has emphasized that LLMs often face knowledge conflicts, where incompatible information can originate from the model’s parametric memory or from retrieved external evidence~\cite{REALM, FiD, lewis2020retrieval, meng2026argus}. A recent survey~\cite{xu2024knowledge} organizes knowledge conflicts into Intra-Context, Inter-Context, and Context-Memory conflicts, and highlights the limitations of existing mitigation approaches. \textbf{For Inter-Context conflicts}, prior work~\cite{chen2022rich} calibrates the model to abstain when retrieved passages contain conflicting answer candidates, rather than forcing a single committed prediction. MADAM-RAG~\cite{wang2025retrieval} mitigates Inter-Context conflicts by delegating each retrieved document to a separate agent for multi-round debate and then synthesizing their deliberations into a final answer. \textbf{For Context-Memory conflicts}, RARR~\cite{gao2023rarr} retrieves evidence after an initial response and performs evidence-grounded revision to better follow the retrieved context. Adacad~\cite{wang2025adacad} estimates the model’s context sensitivity via the distributional discrepancy between decoding with and without context, and uses it to adaptively steer contrastive context-aware decoding. CK-PLUG~\cite{bi2025parameters} detects parameter–context disagreement using a Confidence Gain signal and adjusts token probabilities at decoding time to modulate reliance between parametric knowledge and retrieved evidence.

\paragraph{RLVR for Knowledge Conflicts.} Motivated by recent progress in verifiable-reward training for reasoning models such as DeepSeek-R1~\cite{guo2025deepseek}, 
Reinforcement Learning with Verifiable Rewards (RLVR)~\cite{lambert2024tulu} has become a widely adopted post-training approach to enhance LLM reasoning. More recently, RLVR-based methods have been studied for resolving knowledge conflicts in LLMs~\cite{lin2025knowledgeable, zhang2024trustworthy, zheng2025kcr}. \textbf{For Inter-Context conflicts}, KCR~\cite{zheng2025kcr} pairs two conflicting contexts, extracts structured reasoning paths, and optimizes logic-style rewards to encourage selecting one coherent side rather than blending contradictions. \textbf{For Context-Memory conflicts}, Trustworthy-Alignment~\cite{zhang2024trustworthy} constructs counterfactual contexts and trains the model to prefer the substituted-context answer over its original parametric answer, while Knowledgeable-R1~\cite{lin2025knowledgeable} trains three specialized policies and uses cross-policy advantage comparison to decide when to follow context versus fall back to parametric knowledge under misleading evidence. Although effective, existing RLVR-based methods often treat mixed conflict types uniformly, causing stronger-signal types to dominate optimization and leaving weaker types under-trained. 

\subsection{Imbalance and Exploration in RL for LLMs}
\paragraph{Imbalanced Optimization in Multi-Task RL.}
Recent work shows that jointly optimizing heterogeneous data in RL post-training can bias updates toward certain tasks. Wu et al.~\cite{wu2026imbalanced} systematically analyze multi-task RL post-training and find that some tasks produce much larger gradients and dominate optimization, even though these tasks do not yield larger learning gains. They also observe that absolute advantage tracks gradient magnitude within a task but not across heterogeneous tasks, and they call for future corrections without proposing a concrete method. Several methods address imbalance by reweighting. DISCO~\cite{zhou2025disco} scales rewards according to domain frequency and prompt-level self-consistency, and MT-GRPO~\cite{ramesh2026mtgrpo} adapts task weights to improve worst-task performance. 

\paragraph{Entropy-Guided Exploration in RLVR.}
Token entropy has been widely used to locate critical decisions in reasoning. Wang et al.~\cite{wang2025beyond} find that a small fraction of high-entropy tokens act as forks in reasoning paths, and restrict policy-gradient updates to these tokens. In agentic RL, ARPO~\cite{dong2026arpo} uses an entropy-based adaptive rollout mechanism that branches sampling at high-entropy tool-call steps. ReCon also uses entropy, but for a different purpose: it selects where to branch additional rollouts from failed trajectories. Rather than choosing the token with the highest absolute entropy, ReCon selects the position with the largest local entropy increase among high-entropy candidates. Moreover, branching in ReCon is triggered only for the weaker conflict type, so the additional exploration is driven by type-wise imbalance rather than by entropy alone.

\section{Conclusion}
We study RLVR under mixed knowledge conflicts and identify a previously overlooked failure mode: uniformly optimizing Inter-Context and Context-Memory conflicts produces type-biased updates, where the conflict type with stronger advantage signals dominates while the weaker type remains under-trained.
To address this, we propose \textbf{ReCon}, a conflict-type-aware framework that monitors type-wise mean absolute advantages, detects imbalance, and resamples from the weaker type to rebalance optimization. Specifically, ReCon branches from informative failures by selecting trajectories using uncertainty and closeness signals, and branches at entropy-jump points within those trajectories. 
Experiments on four benchmarks (ConflictBANK, RAMDocs, HotpotQA, and 2WikiMQA) across three model scales (Qwen2.5-3B-Instruct, Qwen2.5-7B- Instruct and Llama-3.1-8B-Instruct) show consistent improvements over strong baselines.

\section*{Limitations}
ReCon relies on reliable conflict-type information to compute type-wise mean absolute advantages and trigger resampling, so its effectiveness may degrade when type signals are missing or noisy. It also introduces extra computation and training-time overhead from imbalance detection, trajectory scoring, and branching-point selection, which can be more noticeable for long reasoning traces and large models. Finally, our empirical validation is centered on knowledge-conflict and multi-hop QA benchmarks under an RLVR setting, and extending ReCon to other conflict taxonomies or retrieval-in-the-loop RAG training remains future work.

\section*{Acknowledgments}
This work was supported by the National Key Research and Development Program of China (Grant No. 2025YFB3003700) and the National Natural Science Foundation of China (Grant No. 62476025).

\newpage
\bibliographystyle{plain}
\bibliography{custom}

\newpage

\appendix


\section{Construction of the Dataset}
\label{app:construction}
The construction procedure is model-specific. Since the subsequent IC/CM assignment depends on memory-state probing, different backbone models may produce different no-context predictions for the same question due to differences in parametric knowledge and reasoning ability. As a result, the constructed datasets are generated separately for each model.

\subsection{Construction from ConflictBANK}
\label{conflictbank}
We build our two conflict settings by re-constructing contexts from ConflictBANK.
In ConflictBANK, each QA instance is paired with one default evidence and three types of conflict evidence: misinformation conflict evidence, temporal conflict evidence, and semantic conflict evidence. 
Following ConflictBANK, misinformation conflicts correspond to inaccurate information, temporal conflicts arise from knowledge changes over time, and semantic conflicts stem from the polysemic nature of language.
In our implementation, these four evidence fields are stored as \texttt{correct} (default evidence), \texttt{fact\_conflict} (misinformation conflict evidence), \texttt{temporal\_conflict} (temporal conflict evidence), and \texttt{semantic\_conflict} (semantic conflict evidence).

\begin{figure}[t]
\centering
\includegraphics[width=0.8\columnwidth]{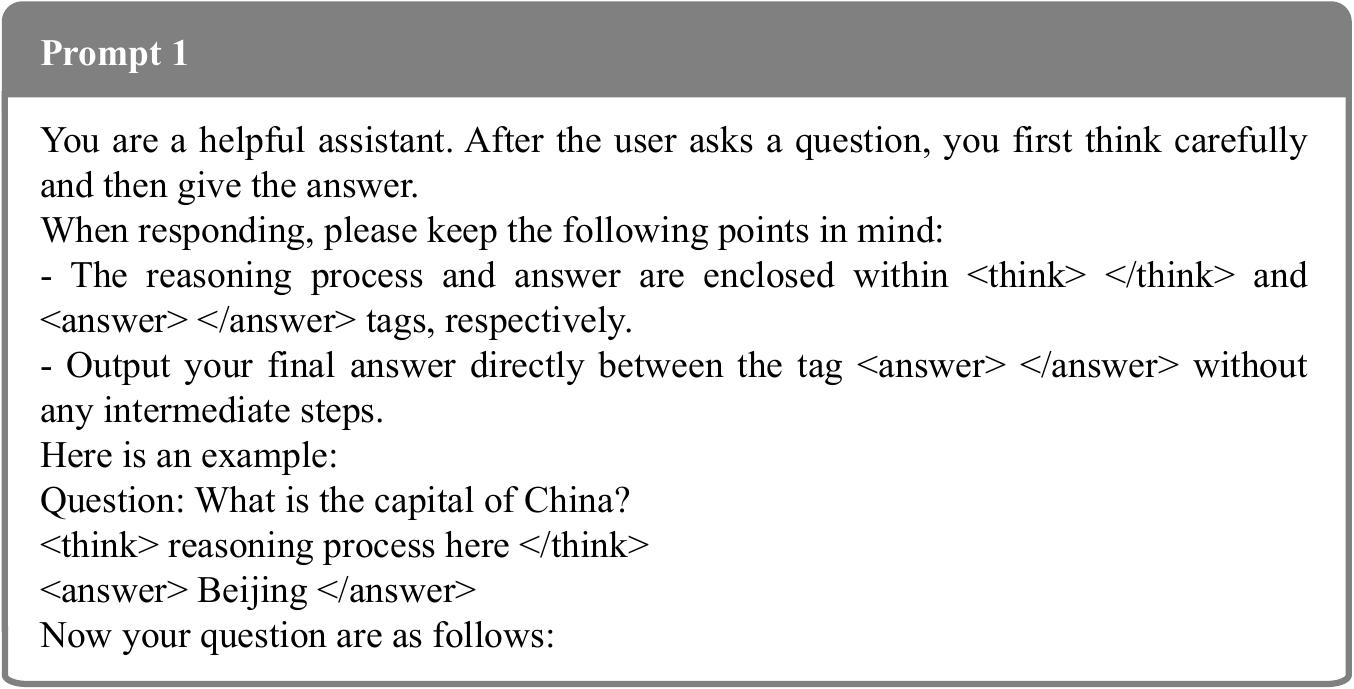}
\caption{Prompt template for Memory-state probing.} 
\label{prompt_m}
\end{figure}

\paragraph{Memory-state probing (no-context answering).}
To explicitly characterize the alignment between the model's parametric memory and the gold answer, we first query the LLM with \emph{only} the question $q$ (without any external documents) and obtain an initial prediction $\hat a$. We use the prompt shown in Figure~\ref{prompt_m}.
We use the correctness of this no-context prediction, i.e., whether $\hat a=a^\star$, to determine the subsequent context construction rule for this instance.

\paragraph{Document sampling and bucket definition.}
For each instance, we sample in total 1--4 documents and concatenate them into a single context $\mathcal{C}$.
Let $c$ be the number of supporting documents sampled from \texttt{correct}, and $w$ be the number of conflicting documents sampled from the union of \texttt{fact\_conflict}, \texttt{temporal\_conflict}, and \texttt{semantic\_conflict}.
We denote each constructed context by a bucket $(c,w)$ and form two settings as follows.

\paragraph{Case A: correct no-context prediction ($\hat a=a^\star$).}
In this case, the model's parametric memory agrees with the gold answer. We construct:
\begin{itemize}[itemsep=2pt,topsep=0pt,parsep=0pt,leftmargin=*]
  \item \textbf{Inter-Context (IC).} We include both supporting and conflicting evidence so that contradictions occur \emph{within} the provided context:
  \[
  (c,w)\in\{(1,1),(1,2),(1,3)\}.
  \]
  \item \textbf{Context-Memory (CM).} We provide only conflicting evidence, making the context directly oppose the model's memory:
  \[
  (c,w)\in\{(0,1),(0,2),(0,3)\}.
  \]
\end{itemize}

\paragraph{Case B: incorrect no-context prediction ($\hat a\neq a^\star$).}
In this case, the model's parametric memory tends to the wrong answer.
To ensure that the context can correct this memory error, we \emph{always} include one supporting document (i.e., $c=1$) and sample $w\in\{0,1,2,3\}$ conflicting documents:
\begin{itemize}[itemsep=2pt,topsep=0pt,parsep=0pt,leftmargin=*]
  \item \textbf{Context-Memory (CM).} We provide only supporting evidence to correct the model's memory:
  \[
  (c,w)=(1,0).
  \]
  \item \textbf{Inter-Context (IC).} We add conflicting evidence on top of the supporting one, creating within-context contradictions and testing evidence selection:
  \[
  (c,w)\in\{(1,1),(1,2),(1,3)\}.
  \]
\end{itemize}

\subsection{Construction from HotpotQA}
\label{hotpotqa}
For multi-hop QA benchmarks, we construct mixed-conflict examples based on HotpotQA. Each constructed instance contains five documents. To determine whether an instance should be used for Inter-Context (IC) or Context-Memory (CM) construction, we follow Appendix~\ref{conflictbank} and first perform memory-state probing by querying the LLM with only the question $q$, without any external documents, and obtain a no-context prediction $\hat a$. We then use whether $\hat a = a^\star$ to determine the subsequent construction rule for this instance.

\textbf{Inter-Context (IC).} For questions that the model can already answer correctly without context, we construct within-context conflicts while keeping the model's parametric knowledge aligned with the gold answer. Each IC instance contains five documents: one first-hop supporting document, one second-hop supporting document, one second-hop conflicting document, and two irrelevant documents. The second-hop conflicting document is constructed by replacing the final answer entity in the original second-hop supporting document with a counterfactual entity, while preserving the surrounding context and sentence structure as much as possible. Similar counterfactual construction has also been explored in prior work on multi-hop QA~\cite{app2023counterfactual, app2024cofca}. This yields a document that remains locally natural but supports an incorrect reasoning path. In this way, the context includes both a correct reasoning path and a conflicting alternative, creating contradictions within the provided evidence.

\textbf{Context-Memory (CM).} For questions that the model answers incorrectly without context, we construct contexts that correct the model's parametric memory rather than introduce within-context contradictions. Each CM instance contains five documents: one first-hop supporting document, one second-hop supporting document, and three irrelevant documents. In this setting, the context consistently supports the gold answer, while the conflict arises between the model's incorrect parametric knowledge and the contextual evidence.

Following this construction, IC samples evaluate the model's ability to resolve contradictions among retrieved documents, whereas CM samples evaluate whether the model can revise incorrect parametric knowledge using supporting context. This allows us to study mixed knowledge conflicts in multi-hop QA under a unified document format.

\begin{figure}[t]
\centering
\includegraphics[width=0.8\columnwidth]{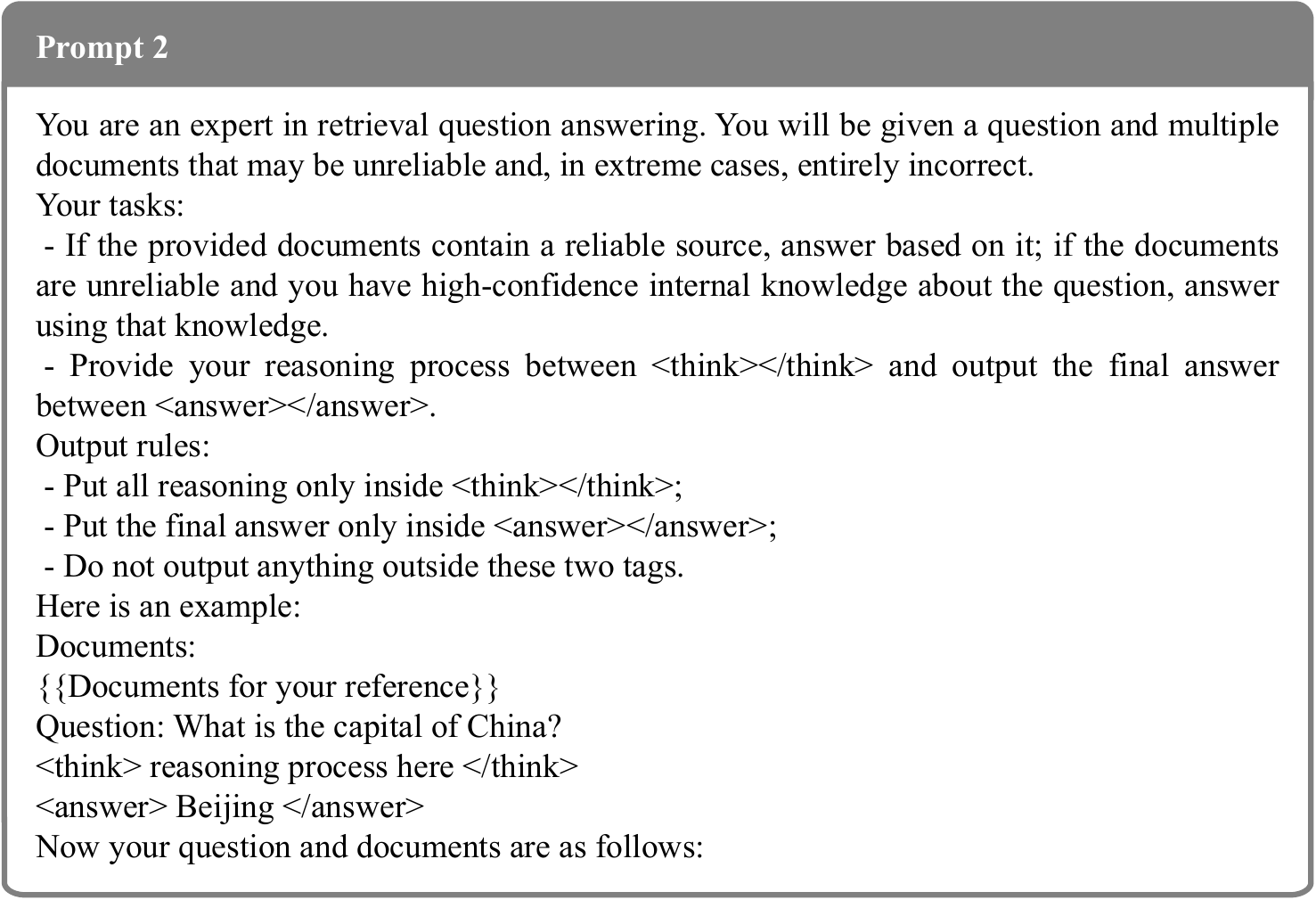}
\caption{Prompt template for ReCon.} 
\label{prompt}
\end{figure}

\section{Evaluation Metric}
\label{app:micro_f1}

We report \textbf{micro-F1} as the main evaluation metric for ConflictBANK and RAMDocs. For a prediction $\hat{y}$ and a gold answer $y^\star$, F1 is computed based on the token-level overlap between the two answers. Specifically, let $P$, $R$, and $F1$ denote precision, recall, and F1, respectively:
\[
P = \frac{|\hat{y} \cap y^\star|}{|\hat{y}|}, \qquad
R = \frac{|\hat{y} \cap y^\star|}{|y^\star|}, \qquad
F1 = \frac{2PR}{P+R}.
\]

For the \textbf{IC} and \textbf{CM} columns, we compute F1 separately on the Inter-Context and Context-Memory subsets. In contrast, the \textbf{ALL} score is computed as \emph{micro-F1} over the union of all IC and CM examples, rather than as the arithmetic mean of the two subset-level scores. Formally, let $\mathcal{D}_{IC}$ and $\mathcal{D}_{CM}$ denote the IC and CM subsets. Then
\[
\text{ALL} = \text{micro-F1}(\mathcal{D}_{IC} \cup \mathcal{D}_{CM}).
\]

\section{Prompt Template}
\label{prompttmp}
We adopt a single prompt template throughout training and evaluation. 
The full template is shown in Figure~\ref{prompt}.

\section{Details of LLMs}
The details of selected LLMs are as follows: 

\textbf{Llama3}~\cite{grattafiori2024llama} is a collection of pre-trained and instruction-tuned LLMs in 8 and 70B sizes. The instruction-tuned LLMs, called Llama-3.1-Instruct, are optimized for dialogue use cases. We selected \textbf{Llama-3.1-8B-Instruct}. 

\textbf{Qwen2.5}~\cite{team2024qwen2} is a series of Qwen large language models, including both base and instruction-tuned versions, spanning a parameter scale from 0.5 billion to 72 billion. We selected \textbf{Qwen2.5-3B-Instruct} and \textbf{Qwen2.5-7B-Instruct}. 

\section{Baseline Implementation Details}
\label{app:baselines}

All baselines, including Trustworthy-Alignment~\citep{TODO_ta}, KCR~\citep{TODO_kcr}, and Knowledgeable-R1~\citep{TODO_kr1}, are retrained from scratch on the same mixed IC--CM training data, backbone models, and evaluation splits as ReCon. We do not use their original checkpoints, training data, or reported results. Below we describe the training configuration of each baseline and how it is adapted to IC and CM examples.

\paragraph{Trustworthy-Alignment.}
We keep its original PPO-based optimization and context-following reward. The actor and critic learning rates are $9.65\times10^{-6}$ and $1\times10^{-6}$, respectively. We use a global batch size of 8, policy and value clipping coefficients of 0.2, $\gamma=1.0$, $\lambda=0.95$, and 10 PPO epochs per batch. The reward coefficients follow the original setting: $\alpha_0=3$, $\alpha_1=1$, $\alpha_2=0.05$, and $\alpha_3=2$, which correspond to the positive/negative answer reward, the unmatched-answer penalty, the KL penalty, and the generation-collapse penalty, respectively. The rollout temperature is linearly annealed from 2.0 to approximately 0, and the maximum prompt and response lengths are 2048 and 512. To adapt the reward to our data, we define the positive and negative answers according to the conflict type. For CM examples, the answer supported by the retrieved context is positive, and the conflicting parametric-memory answer is negative. For IC examples, the answer supported by the reliable context is positive, and the answer supported by the conflicting context is negative. Each response receives a positive, negative, or unmatched reward depending on which answer it contains.

\paragraph{KCR.}
We keep its original GRPO-based optimization and reasoning-path-based reward. We use a learning rate of $2\times10^{-6}$, Adam with $\beta_1=0.9$ and $\beta_2=0.99$, a weight decay of 0.1, a warmup ratio of 0.1, a cosine learning-rate schedule, eight generations per prompt, a maximum gradient norm of 0.1, and 10 training epochs. The maximum sequence length and completion length are 2048 and 512. The logic-coherence, logic-consistency, answer-correctness, and format rewards are weighted by 2, 1, 2, and 1, respectively. Reference and generated reasoning paths are encoded with \texttt{bge-large-en-v1.5}~\citep{xiao2024cpack}. For IC examples, evidence from the reliable context defines the correct side, and evidence from the conflicting context defines the conflicting side. For CM examples, evidence from the retrieved context defines the correct side, while the parametric-memory answer and its supporting evidence define the conflicting side. Based on these definitions, KCR rewards reasoning coherence, reasoning--answer consistency, and answer correctness.

\paragraph{Knowledgeable-R1.}
We keep its original joint optimization over the PK, CK, and RPK objectives. We use a learning rate of $1\times10^{-6}$, a global batch size of 128, a rollout temperature of 1.0, 16 PK and 16 CK trajectories per example, maximum prompt and response lengths of 2048, and 10 training epochs. The PK, CK, and RPK objectives are equally weighted, and the adaptive coefficient $\beta$ is constrained to $[0.01, 1]$. For IC examples, the context input contains both reliable and conflicting documents. For CM examples, it contains the retrieved context that conflicts with the model's parametric knowledge. In both cases, the query-only branch contains only the question, and all trajectories are evaluated against the same ground-truth answer.

\section{Training Cost Analysis}
\label{app:cost}

We compare the training cost of ReCon and GRPO on ConflictBANK with Qwen2.5-7B-Instruct. 
GRPO requires 7.3 hours of training, while ReCon takes 8.6 hours, a 17.8\% overhead, comparable to or smaller than the overhead of related GRPO-based methods such as Knowledgeable-R1 (~24\% over their GRPO baseline).
This extra cost is accompanied by a clear performance gain: ReCon improves the overall F1 score from 76.59\% to 82.75\%, yielding a 6.16-point absolute improvement over GRPO. The additional overhead mainly comes from the extra rollout stage triggered after imbalance detection. Importantly, the gains are not due to uniformly increasing rollouts. As shown in our resampling ablation in Table~\ref{tab:ablation}, uniformly drawing additional rollouts from scratch (\textit{w scratch}) yields much smaller gains than ReCon's targeted resampling, confirming that the extra budget is effective only when focused on the right cases.

\section{Sensitivity Analysis}
\label{app:param}

We report additional sensitivity studies on three key hyperparameters in ReCon: the failure buffer window size $W$, the uncertainty--closeness weighting factor $\alpha_e$, and the trigger threshold $\epsilon_{\text{stop}}$.

\begin{table}[t]
\centering
\small
\caption{Sensitivity to the failure buffer window $W$ on ConflictBANK using Qwen2.5-7B-Instruct. Here, $W$ denotes the number of recent batches whose failed samples are retained in the buffer for resampling. }
\begin{tabular}{lccccc}
\toprule
\textbf{$W$} & 1 & 2 & 4 & 8 & 16 \\
\midrule
\textbf{F1} & 81.26 & 82.06 & \textbf{82.75} & 80.92 & 79.78 \\
\bottomrule
\end{tabular}
\label{tab:w_ablation}
\end{table}

\subsection{Sensitivity to the Failure Buffer Window Size}
\label{app:buffer_window}

We first analyze the sensitivity of ReCon to the failure buffer window size $W$. ReCon maintains a failure buffer for the weaker conflict type and stores failed samples for later resampling. Here, $W$ denotes the number of recent batches whose failed samples are retained in the buffer. It therefore determines the temporal range from which resampling candidates are drawn.

As shown in Table~\ref{tab:w_ablation}, ReCon remains relatively stable across a range of $W$ values, suggesting that the method is not overly sensitive to the exact choice of this window. When $W$ is too small, the buffer contains failures from only a few recent batches, which limits candidate diversity and may cause resampling to focus too narrowly on a small subset of instances. In contrast, when $W$ is too large, the buffer may retain stale failures that no longer reflect the current optimization state of the policy, making resampling less targeted.

The best performance is achieved at a moderate window size, indicating that effective resampling should balance candidate diversity and recency. Overall, these results show that ReCon is robust to the choice of $W$ and benefits from retaining failed samples from a reasonably sized range of recent batches.

\begin{table}[t]
\centering
\small
\caption{Sensitivity to the uncertainty--closeness weight $\alpha_e$ on ConflictBANK using Qwen2.5-7B-Instruct.}
\begin{tabular}{lcccccc}
\toprule
\textbf{$\alpha_e$} & 0 & 0.2 & 0.4 & 0.6 & 0.8 & 1.0 \\
\midrule
\textbf{F1} & 80.92 & 81.38 & \textbf{82.75} & 80.26 & 78.96 & 78.00 \\
\bottomrule
\end{tabular}
\label{tab:alpha_ablation}
\end{table}

\subsection{Sensitivity to the Uncertainty-Closeness Weight}
\label{app:alpha}

We next analyze the effect of the weighting factor $\alpha_e$, which balances the entropy-based uncertainty signal and the preference-based closeness signal in trajectory selection. Larger $\alpha_e$ assigns more weight to uncertainty, whereas smaller $\alpha_e$ places more emphasis on closeness.

As shown in Table~\ref{tab:alpha_ablation}, ReCon achieves the best performance at a moderate value of $\alpha_e=0.4$. This result suggests that effective trajectory selection should remain primarily guided by closeness, while still incorporating a moderate amount of uncertainty information. When $\alpha_e$ is too small, the model relies almost entirely on closeness, which is already a strong signal but does not fully exploit potentially useful uncertainty cues. In contrast, as $\alpha_e$ increases, the method places growing emphasis on uncertainty, and performance drops noticeably when uncertainty becomes dominant.

This trend indicates that closeness is the more reliable signal for identifying correctable failed trajectories, while uncertainty is better viewed as a complementary cue rather than the primary selection criterion. Overall, the results support a closeness-dominant but balanced design, where a moderate contribution from uncertainty yields the strongest performance.

\begin{table}[t]
\centering
\small
\caption{Sensitivity to the trigger threshold $\epsilon_{\text{stop}}$ on ConflictBANK using Qwen2.5-7B-Instruct. Trigger rate denotes the percentage of training steps in which resampling is activated.}
\begin{tabular}{lccccc}
\toprule
\textbf{$\epsilon_{\text{stop}}$} & $3\times10^{-3}$ & $4\times10^{-3}$ & $5\times10^{-3}$ & $6\times10^{-3}$ & $7\times10^{-3}$ \\
\midrule
\textbf{F1} & 82.46 & 82.37 & \textbf{82.75} & 82.11 & 81.89 \\
\textbf{Trigger Rate (\%)} & 90.68 & 90.24 & 89.65 & 89.11 & 88.75 \\
\bottomrule
\end{tabular}
\label{tab:eps_ablation}
\end{table}

\subsection{Sensitivity to the Trigger Threshold}
\label{app:trigger}

We further analyze the sensitivity of ReCon to the trigger threshold $\epsilon_{\text{stop}}$, which determines how pronounced the type-wise imbalance must be before resampling is activated. A larger $\epsilon_{\text{stop}}$ makes resampling more conservative, whereas a smaller value activates it more readily. We use $\epsilon_{\text{stop}}=5\times10^{-3}$ in all main experiments.

As shown in Table~\ref{tab:eps_ablation}, ReCon remains stable across the tested range. The F1 score varies by at most 0.86 points, and the resampling trigger rate varies by only 1.93 percentage points. The default value $\epsilon_{\text{stop}}=5\times10^{-3}$ achieves the best performance. As expected, increasing the threshold slightly reduces the trigger rate, since resampling is activated only when the imbalance between conflict types becomes more pronounced.

Overall, these results show that ReCon does not require careful tuning of $\epsilon_{\text{stop}}$. Moderate changes to the threshold only marginally affect how often resampling is triggered and lead to small differences in final performance.

\section{Multi-Seed Stability}
\label{app:seed}

\begin{table}[t]
\centering
\small
\caption{Multi-seed results on ConflictBANK. We report the mean and sample standard deviation over three random seeds (42, 1024, and 65535). Best results for each backbone are in bold.}
\begin{tabular}{llccc}
\toprule
\textbf{Backbone} & \textbf{Method} & \textbf{IC} & \textbf{CM} & \textbf{ALL} \\
\midrule
\multirow{2}{*}{Qwen2.5-7B-Instruct} & KCR & 82.44 $\pm$ 0.80 & 73.56 $\pm$ 1.20 & 78.54 $\pm$ 1.03 \\
 & ReCon & \textbf{82.64} $\pm$ 1.15 & \textbf{82.70} $\pm$ 0.78 & \textbf{82.60} $\pm$ 0.68 \\
\midrule
\multirow{2}{*}{LLaMA-3.1-8B-Instruct} & Knowledgeable-R1 & 67.10 $\pm$ 1.11 & \textbf{77.12} $\pm$ 1.04 & 72.22 $\pm$ 0.86 \\
 & ReCon & \textbf{79.79} $\pm$ 0.87 & 76.91 $\pm$ 0.93 & \textbf{77.53} $\pm$ 0.53 \\
\bottomrule
\end{tabular}
\label{tab:seed}
\end{table}

To assess whether the improvements of ReCon are stable across random seeds, we rerun the two closest type-level comparisons on ConflictBANK with three random seeds (42, 1024, and 65535): KCR versus ReCon with Qwen2.5-7B-Instruct, and Knowledgeable-R1 versus ReCon with LLaMA-3.1-8B-Instruct. KCR and Knowledgeable-R1 specialize in IC and CM conflicts, respectively, and are the strongest baselines on their favored conflict types. In the main results, these specialized baselines show only marginal advantages on their favored types, whereas the type-level differences in the other comparisons are substantially larger. These two settings are therefore the most informative for testing whether the observed type-level differences are reliable. Table~\ref{tab:seed} reports the mean and sample standard deviation over the three seeds. Since the main results in Table~\ref{tab:main_results} are obtained from a single run, the mean values here differ slightly from those reported in the main text. On their favored conflict types, the specialized baselines and ReCon perform comparably, with differences well within one standard deviation. On the other conflict type, however, ReCon outperforms them by a large margin that far exceeds the seed-level variation, leading to higher overall scores on both backbones. Overall, these results indicate that the advantage of ReCon is not an artifact of a single favorable run.

\section{Compute-Matched Comparison}
\label{app:compute_matched}

\begin{table}[t]
\centering
\small
\caption{Compute-matched comparison on ConflictBANK using Qwen2.5-7B-Instruct. For a batch of $B$ prompts, all methods use at most $8B$ rollouts per training step.}
\begin{tabular}{llccc}
\toprule
\textbf{Method} & \textbf{Rollout budget} & \textbf{IC} & \textbf{CM} & \textbf{ALL} \\
\midrule
GRPO & $8B$ full & 78.43 & 74.15 & 76.59 \\
KCR & $8B$ full & \textbf{82.20} & 74.60 & 79.33 \\
Knowledgeable-R1 & $4B$ PK + $4B$ CK & 73.21 & 76.76 & 74.95 \\
ReCon & $5B$ initial + $\leq 3B$ branched & 81.76 & \textbf{82.48} & \textbf{82.43} \\
\bottomrule
\end{tabular}
\label{tab:compute_matched}
\end{table}

A natural concern is whether the gains of ReCon simply come from the additional rollouts it allocates. To examine this, we compare all methods under a unified budget of at most eight rollouts per prompt. For a batch of $B$ prompts, GRPO and KCR generate $8B$ rollouts per step, so their results in the main experiments are directly comparable. ReCon generates $5B$ initial rollouts and at most $3B$ branched rollouts, so its total budget does not exceed $8B$. Knowledgeable-R1 originally samples 32 trajectories per example; we rerun it with four PK and four CK trajectories while keeping its original objectives and optimization settings. Since ReCon uses a different rollout configuration from the main experiments, its results here differ slightly from those in Table~\ref{tab:main_results}. As shown in Table~\ref{tab:compute_matched}, ReCon achieves the best CM and overall scores under the same maximum budget, while KCR remains slightly better on IC. This indicates that the effectiveness of ReCon cannot be attributed simply to using more rollouts.




\end{document}